\documentclass[runningheads]{llncs}
\usepackage{graphicx}
\usepackage{amsmath,amssymb} 
\usepackage{cite}
\usepackage{booktabs}
\usepackage{hyperref}
\usepackage{tabularx} 
\usepackage{multirow}
\usepackage{longtable}
\usepackage{subcaption}
\usepackage[font={small}]{caption}
\usepackage{pifont}
\usepackage{wrapfig}

\newcommand{\cmark}{\ding{51}}%
\newcommand{\xmark}{\ding{55}}%

\newcommand\blfootnote[1]{%
  \begingroup
  \renewcommand\thefootnote{}\footnote{#1}%
  \addtocounter{footnote}{-1}%
  \endgroup
}

\begin{document}
\title{Improving Deep Visual Representation for Person Re-identification by Global and Local Image-language Association} 

\titlerunning{Re-ID by Global and Local Image-language Association}
%
\author{Dapeng Chen\inst{1} \and
Hongsheng Li$^{\dagger}$\inst{1}  \and Xihui Liu \inst{1} \and Yantao Shen\inst{1} \and Jing Shao \inst{2} 
\and Zejian Yuan\inst{3} \and Xiaogang Wang\inst{1}}
%
\authorrunning{D. Chen et al.}
%

\institute{$^{1}$CUHK-SenseTime Joint Lab, The Chinese University of Hong Kong\\ $^{2}$SenseTime Research ~~~~ $^{3}$Xi'an Jiaotong University \\
\email{\{dpchen, hsli, xhliu, ytshen, xgwang\}@ee.cuhk.edu.hk}\\
}
\maketitle              
\begin{abstract}
Person re-identification is an important task that requires learning discriminative visual features for distinguishing different person identities. Diverse auxiliary information has been utilized to improve the visual feature learning. In this paper, we propose to exploit natural language description as additional training supervisions for effective visual features. Compared with other auxiliary information, language can describe a specific person from more compact and semantic visual aspects, thus is complementary to the pixel-level image data. Our method not only learns better global visual feature with the supervision of the overall description but also enforces semantic consistencies between local visual and linguistic features, which is achieved by building global and local image-language associations. The global image-language association is established
according to the identity labels, while the local association is based upon the implicit correspondences between image regions and noun phrases. Extensive experiments demonstrate the effectiveness of employing language as training supervisions with the two association schemes. Our method achieves state-of-the-art performance without utilizing any auxiliary information during testing and shows better performance than other joint embedding methods for the image-language association. \blfootnote{$^{\dagger}$ Hongsheng Li is the corresponding author.}
\end{abstract}

\section{Introduction}

Person re-identification (re-ID) is a critical task in intelligent video surveillance, aiming to associate the same people across different cameras. Encouraged by the remarkable success of deep Convolutional Neural Network (CNN) in image classification \cite{NIPS2012_Alexnet}, the re-ID community has made great process by developing various networks, yielding quite effective visual representations\cite{ahmed2015improved, varior2016gated, wang2016joint, liu2016spatio, Li_2014_CVPR, Li_Danwei_2017_CVPR, Chen_2017_CVPR, Zhousan_2017_CVPR, liu2017hydraplus, shen2018deep}. To further boost the identification accuracy, diverse auxiliary information has been incorporated in the deep neural networks, such as the camera ID information \cite{Lin2017CVPRcamera}, human poses \cite{zhao2017spindle}, person attributes \cite{su2016deep, attribute_liang}, depth maps\cite{barbosa2012re}, and infrared person images \cite{wu2017rgb}. These data are utilized as either the augmented information for an enhanced inter-image similarity estimation\cite{Lin2017CVPRcamera, zhao2017spindle, wu2017rgb} or the training supervisions that can regularize the feature learning process \cite{su2016deep, attribute_liang}. Our work belongs to the latter category and proposes to use language descriptions as training supervisions to improve the person visual features. Compared with other types of auxiliary information, natural language provides a flexible and compact way of describing the salient visual aspects for distinguishing different persons. Previous efforts on language-based person re-ID \cite{Lishuang_2017_CVPR} is about cross-modal image-text retrieval, aiming to search the target image from a gallery set by a text query. Instead, we are interested in how the language can help the image-to-image search when they are only utilized in the training stage. This task is non-trivial because it requires a detailed understanding of the content of images, language, and their cross-modal correspondences.

\begin{figure}[t]
\vspace{-1em}
\centering
\includegraphics[width=0.9\textwidth]{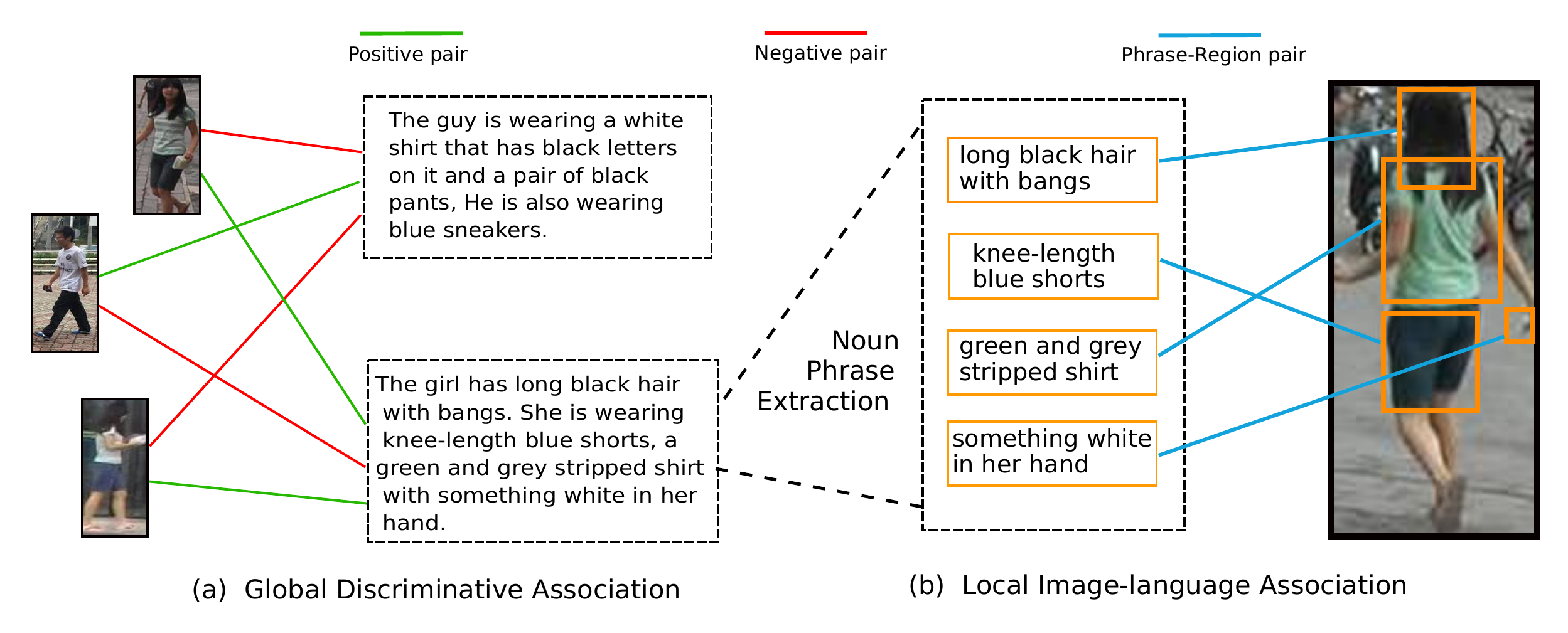}
\vspace{-1em}
\caption{Illustration of global and local image-language association in our framework. The global association is applied to the whole image and the language description, aiming to discriminate the matched image-language pairs from the unmatched ones. The local association aims to model the correspondences between the noun-phrases and images regions. The global and local image-language association is utilized to supervise the learning of person visual features.}  \label{illustration}
\vspace{-1em}
\end{figure}

 To exploit the semantic information conveyed in the language descriptions, we not only need to identify the final image representation but also propose to optimize the global and local association between the intermediate features and linguistic features. The global image-language association is learned from their ID labels. That is, the overall image feature and text feature should have high relevance for the same person, and have low relevance when they are from different persons (Fig. \ref{illustration}, left). The local image-language association is based on the implicit correspondences between image regions and noun phrases (Fig . \ref{illustration}, right). As in a coupled image-text pair, a noun phrase in the text usually describes a specific region in the image, thus the phrase feature is more related to some local visual features. We design a deep neural network to automatically associate related phrases and local visual features via the attention mechanism, then aggregate these visual features to reconstruct the phrase. Reasoning such latent and inter-modal correspondence makes the feature embedding interpretable, can be employed as a regularization scheme for feature learning.
 
In summary, our contributions are three-fold: (1) We propose to use language description as training supervisions for learning more discriminative visual representation for person re-ID. This is different from existing text-image embedding methods aiming at cross-modal retrieval. (2) We provide two effective and complementary image-language association schemes, which utilize semantic,  linguistic information to guide the learning of visual features in different granularities. (3) Extensive ablation studies validate the effectiveness and complementarity of the two association schemes. Our method achieves state-of-the-art performance on person re-ID and outperforms conventional cross-modal embedding methods. 
                      
\section{Related Work}

Early works on person re-ID concentrated on either feature extraction \cite{wang2007shape, ma2012bicov, farenzena2010person}  or metric learning \cite{koestinger2012large, chen2015similarity, chen2016similarity, mignon2012pcca, chen2017exemplar}. Recent methods mainly benefit from the advances of CNN architectures \cite{Lishuang_2017_CVPR},  which combine the above two aspects to produce robust and ID-discriminative image representation \cite{Li_2014_CVPR, ahmed2015improved, varior2016gated, wang2016joint, shen2018end, chen2018group}. Our work aims to further improve the deep visual representation by making use of language descriptions as training supervisions.

Diverse auxiliary information has been introduced to improve the visual feature representations for person re-ID. Several works \cite{zhao2017spindle, Su_2017_ICCV, ZhengHLY17} detected person pose landmarks to obtain the human body regions.  They firstly decomposed the feature maps according to the regions,  
then fused them to create the well-aligned feature maps.  Lin \emph{et al.} utilized Camera ID information to assist inter-image similarity estimation \cite{Lin2017CVPRcamera} by keeping consistencies in a camera network. Also, different types of sensors such as depth cameras \cite{barbosa2012re}, or infrared \cite{wu2017rgb} cameras have been employed in person re-ID to generate more reliable visual representations. For these methods, the auxiliary information is used in both training and testing stage, requiring an additional model or data acquisition device for algorithm deployment.  Differently, human attributes usually serve as a kind of training supervisions. For example, Su \emph{et al.} \cite{su2016deep} learned a semi-supervised discriminative model to predict the  binary attribute feature for re-ID. Lin \emph{et al.} \cite{attribute_liang} improved the interpretability of the intermediate feature maps by jointly optimizing the identification loss and attribute classification loss.  Although attributes proves helpful for feature learning, they are quite difficult to obtain as people need to remember tens of attribute labels for annotations. They are also less flexible to describe diverse variations in human appearance.

Associating image and language helps establish correspondences for their inter-relations. It has attracted great attention  in recent years because of its wide applications in image captioning \cite{Karpathy_2017_PAMI, vinyals2015show, xu2015show, Mindeye,liu2018show}, visual QA \cite{antol2015vqa,li2018visual, Johnson_2016_CVPR}, and text-image retrieval \cite{frome2013devise, reed2016learning}. These cross-modal associations can be modeled by either generative methods or discriminative methods. Generative models utilize probabilistic models to capture the temporal or spatial dependencies within the image or text \cite{mirza2014conditional, vinyals2015show}, and have popular applications like caption generation \cite{vinyals2015show, xu2015show, Rennie_2017_CVPR, Anderson2017up-down, liu2018show} and image generation \cite{pmlr-v48-reed16, reed2016learning}. On the other hand, discriminative models have also been developed for image-text association. Karpathy and Fei-Fei \cite{Karpathy:2014} formulated a bidirectional ranking loss to associate the text and image fragments.  Reed \emph{et al.} \cite{reed2016learning} proposed deep symmetric structured joint embeddings, and enforced the embedding of matched image-text pair should be higher than those of unmatched pairs. Our method combines the merits of both discriminative and generative methods to build image-text association in different granularities, where the language descriptions act as training supervisions to improve visual representation.

\begin{figure}[t]
\vspace{-1em}
\centering
\includegraphics[width=0.92\textwidth]{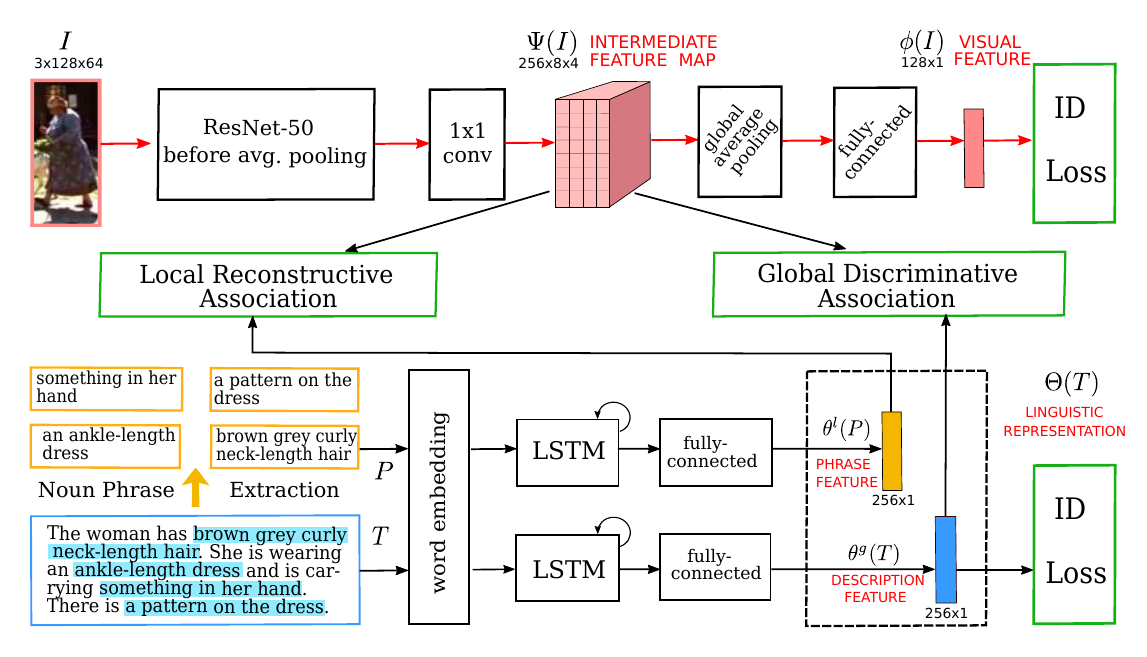}
\vspace{-1em}
\caption{Overall framework of our proposed approach. We employ the ResNet-50 as the backbone architecture. The produced intermediate feature $\Psi(I)$ is associated to the description feature $\theta^{g}(T)$ and the phrase feature $\theta^{l}(P)$ by global discriminative association and local reconstructive association, respectively.}  \label{Overall_framework}
\vspace{-1em}
\end{figure}

\section{Our Approach}

To improve the visual representation for person re-ID with deep neural networks, we aim to exploit language descriptions of person images as the training supervisions in addition to the original ID labels. The visual representations are not only required to be discriminative for different persons but also need to keep consistencies with the linguistic representations. We, therefore,  propose the global and local image-language association schemes. The global visual feature of one person should be more relevant to the language description features of the same person than those of a different person. Unlike existing cross-modal joint embedding methods, we do not require the visual and linguistic features to be mapped to a unified embedding space. Furthermore, based on the assumption that the image and language are spatially decomposable and temporally decomposable, we also try to find the mutual correspondences between the features of the image regions and the noun-phrases. The overall framework is illustrated in Fig. 
\ref{Overall_framework}.

\subsection{Visual and Linguistic Representation}

Given a dataset $\mathcal{D}=\{(I_{n}, T_{n}, l_{n})\}_{n=1}^{N}$ containing $N$ tuples, each tuple has an image $I$, a text description $T$, and an ID label $l$. To improve the learned visual feature $\phi(I)$, we build global and local correspondences between the intermediate visual feature maps $\Psi(I)$ and linguistic representation $\Theta(T)$. 

\vspace{0.5em}

\noindent \textbf{The visual representation.}  The visual feature $\phi(I)$ and the intermediate feature map $\Psi(I)$ are obtained from standard convolutional neural network (CNN), which takes ResNet-50 as the backbone network. $\Psi(I)$ is the feature map obtained with the $1\!\times\!1$ convolution over the last residual-block. Suppose the $\Psi(I)$ has $K$ bins, the feature vector at the $k$th bin is denoted by $\psi_{k}(I)$, then $\Psi(I)$ can be represented as $\Psi(I) =\{ \psi_{k}(I) \}_{k=1}^{K}$.  The objective visual feature vector $\phi(I)$ is linear projection from the average feature map $\bar{\psi}(I)=\frac{1}{K}\sum_{k=1}^{K}\psi_{k}(I)$:
\begin{equation}\small
\phi(I)=f_{\phi}(\Psi(I)) = \bold{W}_{\phi}\bar{\psi}(I) + \bold{b}_{\phi}. \label{final_map}
\end{equation}
We employ the ID loss over $\phi(I)$, aiming to make it distinctive for different persons. Specifically, given $N$ images belonging to $I$ persons, the ID loss is the average negative log-likelihood of the feature maps being correctly classified to its ID:
\begin{equation}\small
 \mathcal{L}_{I} = -\frac{1}{N} \sum_{n=1}^{N}\sum_{i=1}^{I}y_{i,n} \log \left(\frac{\exp(\bold{w}_{i}^{\top} \phi(I_{n})   )}{\sum_{j=1}^{I}\exp(\bold{w}_{j}^{\top}   \phi(I_{n}))} \right), \label{Eqn:ID-loss}
\end{equation}
where $y_{i,n}$ is the index label with $y_{i,n}=1$ if the $n$th image $I_{n}$ belongs to the $i$th person and $y_{i,n} = 0$ otherwise.  $\bold{w}_{i}$ are the classifier parameters associated with the $i$th person over the visual feature vectors.

\vspace{0.5em}

\begin{figure}[t]
 \vspace{-1em}
 \centering
\includegraphics[width=0.85\textwidth]{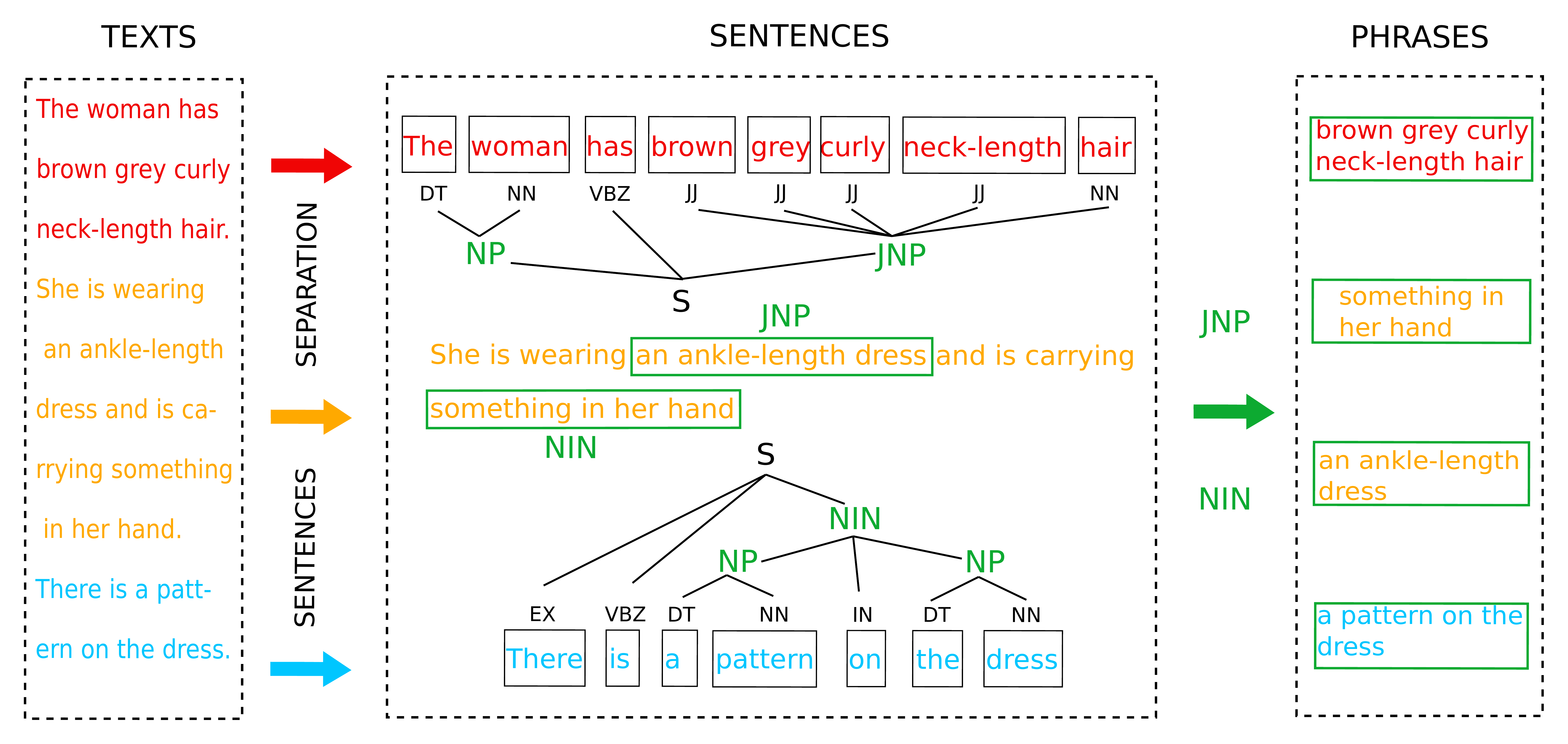}
 \vspace{-1em}
 \caption{The flowchart of extracting interested noun phrases from the text. We perform  word-level tokenization and part-of-speech tagging, then extract noun phrases by chunking. As not all the phrases can have discriminative information, we are interested in two kinds of the phrases: (1) the noun phrase with adjectives (JJ), defined as JNP (2) the noun phrase consists of multiple nouns joined by preposition (IN).}  \label{fig:noun-phrase-extraction} 
\vspace{-1em}
\end{figure}

\noindent \textbf{The linguistic representation.} $\Theta(T)$ contains two types of feature vectors as shown in Fig. \ref{Overall_framework}. One is the global description feature $\theta^{g}(T)$ mapped from the whole text, the other is the local phrase feature $\theta^{l}(P) $ that encodes a distinctive noun phrase $P$ cropped from the text $T$. The noun-phrase extraction procedure is demonstrated in Fig. \ref{fig:noun-phrase-extraction}, and the obtained phrases in $T$ form the set $\mathcal{P}(T)$. Each word in text $T$ or phrase $P$ is firstly represented as a $D$ dimensional one-hot vector, denoted by $\bold{o}_{m}\in \mathbb{R}^{D}$ for the $m$th word and $D$ is the vocabulary size. Then the one-hot vector is projected to a word embedding: $\bold{e}_{m} = \bold{W}_{e} \bold{o}_{m}$.

Based on the embedding, we feed either a whole description or a short phrase to a long short-term memory network 
(LSTM) word by word, which has the following updating procedure: $\bold{h}_{m+1} = \text{LSTM}(\bold{e}_{m}, \bold{h}_{m})$. The LSTM unit takes the current word embedding $\bold{e}_{m}$ and hidden state $\bold{h}_{m}$ as inputs,  and outputs the hidden state of the next step $\bold{e}_{m+1}$.
The hidden states at the final time step are effective summarization of the description $T$ or phrase $P$, obtaining the description feature $\theta^{g}(T)$ or  the phrase feature $\theta^{l}(P)$ by:
\begin{equation}\small
  \theta^{g}(T) = \bold{W}_{g} \bold{h}_{F}(T) + \bold{b}_{g}, \qquad \theta^{l}(P) = \bold{W}_{l}\bold{h}_{F}(P) + \bold{b}_{l},
\end{equation}
where $\bold{h}_{F}(T)$ and $\bold{h}_{F}(P)$ are the final hidden states for text $T$ and phrase $P$, respectively. Because $T$ describes abundant person characteristics throughout the body, $\theta^{g}(T)$ could describe a specific person. We therefore impose another ID loss to make $\theta^{g}(T)$ be separable for different persons,
\begin{equation}\small
\mathcal{L}_{T} = - \frac{1}{N}\sum_{n=1}^{N} \sum_{i=1}^{I} y_{i,n}\log\left(\frac{\exp(\bold{v}_{i}^{\top} \theta^{g}(T_{n}))}{\sum_{j=1}^{I}\exp(\bold{v}_{j}^{\top} \theta^{g}(T_{n}))} \right),
\end{equation}
where $y_{i,n}$ is defined in the same way with the one in Eqn. (\ref{Eqn:ID-loss}), and $\bold{v}_{i}$ indicates the classifier parameters associated with the $i$th person over the description feature.
 
\subsection{Global Discriminative Image-language Association}

The ID losses in the previous section only enforce the visual and linguistic feature be discriminative within each modality but do not establish image-language correspondences to enhance the visual feature. As the global description is usually related to multiple and diverse regions in the image,  $\theta^{g}(T)$ can be associated to $\bar{\psi}(I)$ (Eqn. (\ref{final_map})) in a discriminative fashion. Specifically, $\bar{\psi}(I)$ and $\theta^{g}(T)$ firstly form a joint representation  $\varphi(I, T)$:
\begin{equation} \small
   \varphi(I, T) = \big(\bar{\psi}(I) - \theta^{g}(T) \big) \circ \big(\bar{\psi}(I) - \theta^{g}(T) \big),  \label{Eqn:r1}
\end{equation}
where $\circ$ denotes the Hadamard product. The joint representation is then projected into a scalar value within the range $(0, 1)$ by:
\begin{equation} \small
    s(I, T) = \frac{ \exp(\bold{w}_{s}^{\top}\varphi(I, T)+ b_{s})}{1+ \exp(\bold{w}_{s}^{\top}\varphi(I, T)+ b_{s})}. \label{Eqn:r2}
\end{equation}
To build the relevance between $\bar{\psi}(I)$ and $\theta^{g}(T)$, we expect $s(I, T)$ to be 1 when $I$ and $T$ belong to the same person and to be $0$ when they belong to different persons. We thus impose the binary cross-entropy loss over the scores:
\begin{equation}\small
 \mathcal{L}_{dis} = - \frac{1}{\hat{N}}\sum_{i,j} \Big[  l_{i,j}\log\big( s(I_{i}, T_{j}) \big) +   (1-l_{i,j})\log\big(1-s(I_{i}, T_{j})\big) \Big],  
 \label{Eq:loss_dis}
\end{equation}
where $\hat{N}$ is the number of sampled image-text pairs. $l_{i,j} =1$  if $I_{i}$ and $T_{j}$ are describing a same person and $l_{i,j}=0$ otherwise.

\noindent \textbf{Discussion}.  Here, we draw a distinction between the proposed discriminative scheme  and the bi-directional ranking \cite{Karpathy:2014,reed2016learning, Dual-path}, which is formulated by:
\begin{equation}\small
 \mathcal{L}_{rank}= \frac{1}{\hat{N}}\sum_{i,j}\max(0, k_{i,j}- k_{i,i}+ \alpha )+ \max(0, k_{j,i}-k_{i,i} + \alpha),  \label{Eq:loss_rank}
\end{equation}
where $k_{i,j} = \bar{\psi}(I_{i})^{\top}\theta^{g}(T_{j})$. The loss stipulates that the cosine similarity  $k_{i,i}$ for one image-text tuple should be higher than $k_{i,j}$ or $k_{j,i}$ for any $i \!\neq\! j$ by at least a margin of $\alpha$.  We highlight two main differences between the proposed $\mathcal{L}_{dis}$(Eqn. (\ref{Eq:loss_dis})) and $\mathcal{L}_{rank}$: (1) As $\mathcal{L}_{rank}$ is originally applied in the image-text retrieval task,  it associates the image and text description features by simply checking whether they are from the same tuple. Differently, $\mathcal{L}_{dis}$ is based on person ID, which is more reasonable as one description can well correspond to different images of the same person. (2) $\mathcal{L}_{rank}$ estimates the image-text relevance by cosine similarity, requiring $\bar{\psi}(I_{i})$ and $\theta^{g}(T_{j})$ lie in the same feature space. Meanwhile, our scheme employs a projection over the joint representation, being able to capture more complicated correlations between image and text description.

\subsection{Local Reconstructive Image-language Association}\label{Sec:LRA}

A phrase usually only describes one part of an image and could be contained in the descriptions of different persons. For this reason, a phrase is disjoint with the person ID, but can still build correspondences with a certain region in the image it describes. We therefore propose a reconstruction scheme.  That is, the phrase feature $\theta^{l}(P)$ can select relevant feature vectors in visual feature map $\Psi(I_{n})$ if $P \in \mathcal{P}(T_{n})$, and the selected feature vectors are able to reconstruct the phrase $P$ in turn.

\vspace{0.5em}

\begin{figure}[t]
 \begin{center}
\includegraphics[width=0.98\textwidth]{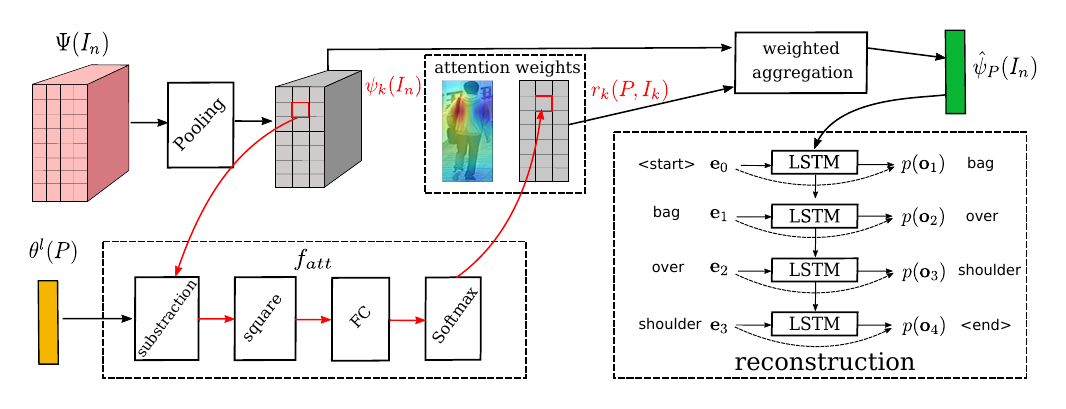}
 \vspace{-1.5em}
 \caption{The network structure for the local reconstructive image-language association. We first use the feature maps $\Psi(I_{n})$ and the phrase feature $\theta^{l}(P)$ to compute the attention weights for the intermediate features at different locations, then perform weighted aggregation to obtain the visual feature $\hat{\psi}_{P}(I_{n})$, and finally employ LSTM to reconstruct $P$ with $\hat{\psi}_{P}(I_{n})$.}  \vspace{-2em} \label{fig:local-reconstruction} 
 \end{center}
\end{figure}

\noindent \textbf{Image feature aggregation.} Suppose $P$ is a phrase that describes a specific region in image $I_{n}$, we aim to estimate a vector $\hat{\psi}_{P}(I_{n})$ that can reflect the features in the region. For this purpose, we compute $\hat{\psi}_{P}(I_{n})$ by weighted aggregation of the feature vectors $\{\psi_{k}(I_{n})\}_{k=1}^{K}$ in the feature map $\Psi(I_{n})$: 
\begin{equation}\small
     \hat{\psi}_{P}(I_{n}) = \sum_{k=1}^{K} r_{k}(P,I_{n})\psi_{k}(I_{n}), \label{Eq:agg_feature}
\end{equation}
where $r_{k}(P, I_{n})$ is the attention weight reflecting the relevance between the phrase $P$ and the feature vector $\psi_{k}(I_{n})$ . It is estimated by an attention function $f_{att}\big(\psi_{k}(I_{n}), \theta^{l}(P) \big)$, which first computes the the unnormalized weight $ \bar{r}_{k}(P, I_{n})$ with a linear projection over the joint representation of $\psi_{k}(I_{n})$ and  $r_{k}(P, I_{n})$:
\begin{equation}\small
      \bar{r}_{k}(P,I_{n}) = \bold{w}_{\bar{r}}^{\top} \big((\psi_{k}(I_{n})- \theta^{l}(P)) \circ (\psi_{k}(I_{n})- \theta^{l}(P))\big)  + b_{\bar{r}},
\end{equation}
then normalizes the values by using a softmax operation over all the $K$ bins:
\begin{equation}\small
   r_{k}(P,I_{n})= \textstyle{\exp(\bar{r}_{k}(P, I_{n})) /\sum_{k=1}^{K} \exp(\bar{r}_{k}(P, I_{n}))}. \label{Eqn:attention_weights}
\end{equation}
In practice, the attention model is easy to overfit with limited training data.  Besides, the spatially adjacent feature maps possibly represent one phrase, they are more reasonable to be merged. For these reasons, we reduce the training burden by average pooling the neighboring feature maps in $\Psi(I_{n})$ before the weighted aggregation, which is also illustrated in Fig. \ref{fig:local-reconstruction}.  

\vspace{0.5em}

\noindent \textbf{Phrase reconstruction.}  To enforce the consistency between the aggregated feature map  $ \hat{\psi}_{P}(I_{n})$ and the input phrase $P$, we build the conditional probability $p(P| \hat{\psi}_{P}(I_{n}))$ to reconstruct $P$ with $ \hat{\psi}_{P}(I_{n})$. Since a phrase has a unbounded length $M$, it is common to apply the chain rule to model the probability over $\{\bold{o}_{1}, \bold{o}_{2}, ..., \bold{o}_{M+1} \}$,
\begin{equation} \small
  \log p(P|\hat{\psi}_{P}(I_{n})) = \sum_{m=0}^{M} \log p\big(\bold{o}_{m+1}| \hat{\psi}_{P}(I_{n}), \hat{\bold{o}}_{0},..., \hat{\bold{o}}_{m}\big),
\end{equation}
where $\bold{o}_{m+1}(m=0,...,M)$ is the random variable over the one-hot vectors of the m-$th$ word, and $\{\hat{\bold{o}}_{0},...,\hat{\bold{o}}_{M+1}\}$ are one-hot vectors of the ground truth words. Among them, $\hat{\bold{o}}_{0},\hat{\bold{o}}_{M+1}$ are the one-hot vectors that designate the start and end of the phrase. Inspired by the task of image caption generation \cite{vinyals2015show, xu2015show}, LSTM is employed to model $ p \big(\bold{o}_{m+1}| \hat{\psi}_{P}(I_{n}), \hat{\bold{o}}_{0},..., \hat{\bold{o}}_{m} \big)$. More specifically, we initially feed $\hat{\psi}_{P}(I_{n})$ to the LSTM, then feed the embedding of the current word to obtain the hidden state of the next word. The next word probability is computed from the hidden state $\bold{h}_{m\!+\!1}$ and the word embedding $\bold{e}_{m}$. The word probability can be formulated as: $p\big( \bold{o}_{m+1}| \hat{\psi}_{P}(I_{n}), \hat{\bold{o}}_{0},..., \hat{\bold{o}}_{m} \big) \propto \exp(\bold{W}_{oh}\bold{h}_{m+1} + \bold{W}_{oe}\bold{e}_{m})$. The reconstruction loss is the sum of the negative log likelihood of the correct word at each step:
\begin{equation} \small
    \mathcal{L}_{rec} = -\frac{1}{N}\sum_{n=1}^{N} \frac{1}{|\mathcal{P}(T_{n})|}\sum_{P\in \mathcal{P}(T_{n})} \log p\big(P| \hat{\psi}_{P}(I_{n})\big). \label{Eqn:rec}
\end{equation}

\subsection{Training and Testing}

The final loss function is a combination of the image ID loss, the text ID loss as well as the discriminative and reconstructive image-language association losses:
\begin{equation} \small
\mathcal{L} = \mathcal{L}_{I}+ \lambda_{T}\mathcal{L}_{T}+ \lambda_{dis} \mathcal{L}_{dis}+\lambda_{rec} \mathcal{L}_{rec}, \label{Eqn:jointloss}
\end{equation}
where $\lambda_{T}, \lambda_{dis}$ and $\lambda_{rec}$ are balancing parameters. For network training, we adopt stochastic gradient descent (SGD) with an initial learning rate of $10^{-2}$, which is further decayed to $10^{-3}$ after the 20th epoch.  We organize the training batch as follows.
The data tuple $(I_{n}, T_{n}, d_{n})$ is firstly transformed to $(I_{n}, T_{n}, \mathcal{P}(T_{n}), d_{n})$.  Each batch contains the samples from 32 randomly selected persons, and each person has two randomly sampled tuples. For global discrimination, we form  $32 \times 4$ positive image-description pairs by exploiting all the intra-tuple and inter-tuple image-description compositions, and sample 6 negative pairs for each image, yielding $64\times 6$ negative pairs, keeping the pos/neg ratio to be 1:3. Meanwhile, the local reconstruction is performed within each tuple.

In testing, only image features are extracted, and no language descriptions are used. The distance between two image features are simply the Euclidean distance, \emph{i.e.,}
\begin{equation}\small
  d_{i,j} = \|\phi(I_{i}) - \phi(I_{j})\|_{2}.
\end{equation}
Person Re-ID is performed by ranking the distances between the probe image and gallery images in ascending order.

\section{Experiments}

We evaluate the proposed approach on three standard person re-ID datasets, whose language annotations can be fully or partially obtained from the CUHK-PEDES dataset \cite{Lishuang_2017_CVPR}. Ablation studies are mainly conducted on Market-1501 \cite{zheng2015scalable} and CUHK-SYSU \cite{journals/corr/XiaoLWLW16}, which are  convenient for extensive evaluation as with fixed training/testing splits. We also report the overall results on Market-1501, CUHK03\cite{Li_2014_CVPR} and CUHK01\cite{li2012human} to compare with the state-of-the-art approaches.

\subsection{Experimental Setup} \label{Sec:Setup}

\noindent \textbf{Datasets and Metrics.} \quad  To verify the utility of language descriptions in person re-ID, we augment four standard person re-ID datasets (Market-1501, CUHK03, CUHK01, and CUHK-SYSU) with language descriptions. The language descriptions are obtained from the CUHK-PEDES  dataset, which is originally developed for cross-modal text-based person search and  contains 40,206 images of 13,003 persons from five existing person re-ID datasets. Since persons in Market-1501 and CUHK03 have many similar samples, only four images of each person in this two datasets have language descriptions. 

Among the four datasets, Market-1501 consists of 32,668 images of 1,501 persons and provides a standard protocol for training and testing.  CUHK03 contains 13,164 images of 1,360 persons. Following \cite{Li_2014_CVPR}, we use 1,260 persons for training and the rest 100 persons for testing. CUHK01 contains 971 persons captured from two views, and each person has two images in each view. 485 persons are randomly selected for training, and the remaining 486 persons are used for testing. CUHK-SYSU is a new dataset used for joint detection and identification. According to separation in CUHK-PEDES, 15,080 images from 5,532 identities are used for training, 8,341 images from 2,900 persons are used for testing with 2,900 query images and 5,441 gallery images. Mean average precision (mAP) and CMC top-1, top-5, top-10 accuracies are adopted as the evaluation metrics. 

\vspace{0.5em}

\noindent \textbf{Implementation details.}  All the person images are resized to 256$\times$128. For data augmentation, random horizontal flipping and random cropping are adopted. We empirically set the dimensions of feature embeddings $\phi(I)$, $\theta^{l}(P)$ and $\theta^{g}(T)$ to be $256$, and set the  balancing parameters $\lambda_{T}=0.1$, $ \lambda_{dis}=1$, $\lambda_{rec}=1$, respectively.  As some images in Market-1501 and CUHK03  do not have language descriptions, we employ the description of the same person (in the same camera if possible) for them to compose the data tuple $(I_{n}, T_{n}, d_{n})$. The ResNet-50 backbone is initialized by the parameters pre-trained on ImageNet\cite{imagenet_cvpr09}.

\begin{table}[h]\vspace{-1em}
\setlength{\tabcolsep}{6pt}
\begin{center}
  \resizebox{0.70\hsize}{0.11\hsize}{
   \begin{tabular}{p{85pt}p{28pt}<{\centering}p{28pt}<{\centering}p{28pt}<{\centering}p{28pt}<{\centering}p{40pt}<{\centering}}
    \toprule
   \multicolumn{1}{c}{\multirow{2}{*}{Methods}}  & \multicolumn{5}{c}{ Training Loss} \\
 \cline{2-6}  
                                &  $\mathcal{L}_{I}$ & $\mathcal{L}_{T}$ & $\mathcal{L}_{dis}$ & $\mathcal{L}_{rec}$  & $\mathcal{L}_{rank}$ \\
   \hline
    \emph{basel.}                       &      \cmark      &      \xmark          &     \xmark          &      \xmark       &       \xmark          \\
    \emph{basel. \!+\! rank}         &      \cmark      &      \cmark          &     \xmark          &     \xmark        &     \cmark     \\  
    \emph{basel. \!+\! GDA}             &      \cmark      &        \cmark          &       \cmark     &  \xmark           &     \xmark            \\   
    \emph{basel. \!+\! LRA}             &      \cmark      &      \xmark          &      \xmark         &       \cmark       &       \xmark          \\
    \emph{proposed}         &      \cmark      &      \cmark          &     \cmark          &        \cmark      &        \xmark         \\
  \bottomrule
   \end{tabular}
   }
 \end{center} \vspace{-1em}
 \caption{The loss configurations for the baseline and other variants.} \label{Tab:config} \vspace{-2em}
\end{table}

\noindent\textbf{Baseline and variants.} The baseline is just the visual CNN that produces the feature map $\phi(I)$, indicated by the red lines in Fig. \ref{Overall_framework}.  We additionally build 4 variants on the baseline for ablation study. The loss configuration of them are displayed in Table. \ref{Tab:config}. Among them, \emph{basel.} only imposes the ID loss to make $\phi(I)$ be separable for different persons. Both \emph{basel.+rank} and \emph{basel.+GDA} additionally impose the ID loss over the global description feature $\theta^{g}(T)$ but have different global image-language association schemes. \emph{basel.+rank} employs the $\mathcal{L}_{rank}$ in Eqn.(\ref{Eq:loss_rank}), while \emph{basel.+GDA} utilizes the proposed $\mathcal{L}_{dis}$ in Eqn.(\ref{Eq:loss_dis}). The variant \emph{basel.+LRA}
employs the reconstruction loss $\mathcal{L}_{rec}$ in Eqn.(\ref{Eqn:rec}) to build the local association between the aggregated feature vector $\hat{\psi}_{P}(I_{n})$ and the phrase feature $\theta^{l}(P)$. Our proposed method takes advantages of both global and local image-language association schemes.

 \begin{table}[t]
 \vspace{-1em}
\setlength{\tabcolsep}{6pt}
 \begin{center}
  \resizebox{0.98\hsize}{0.12\hsize}{
   \begin{tabular}{p{80pt} p{23pt}<{\centering}p{23pt}<{\centering}p{23pt}<{\centering}p{28pt}<{\centering}   					 p{1pt}<{\centering}			
                                        p{23pt}<{\centering}p{23pt}<{\centering}p{23pt}<{\centering}p{28pt}<{\centering}   
   }

    \toprule
   \multicolumn{1}{c}{\multirow{2}{*}{Methods}}  & \multicolumn{4}{c}{Market-1501} &  & \multicolumn{4}{c}{CUHK-SYSU}  \\
 \cline{2-5}   \cline{7-10}
                                &  mAP & top-1 & top-5 & top-10  &   & mAP  & top-1 & top-5 & top-10  \\
   \hline
  \emph{basel.}                                         & 74.4  & 89.2  & 95.5   & 96.9   &    & 85.8  &  87.3  &  93.7  &  95.1 \\
  $\emph{basel.\!+\! rank}^{1}$\cite{Karpathy:2014}  & 75.5  & 88.5  & 95.9   & 97.5   &    & 87.0  &  88.0  &  94.5  &  95.9 \\
  $\emph{basel.\!+\! rank}^{2}$\cite{Dual-path}         & 77.7  & 90.5  & 96.1   & 97.6   &    & 88.8  &  90.2  &  95.5  &  96.8 \\
  \emph{basel. \!+\! GDA}                               & 80.0  & 91.5  & 96.4   & 98.0   &    & 90.2  &  91.0  &  96.2  &  97.5 \\ 
  \emph{basel. \!+\! LRA}                               & 79.6  & 91.6  & 96.7   & 97.9   &    & 89.7   &  90.7  &  96.0   &  97.4 \\ 
  \emph{proposed}    & \textbf{81.8}    & \textbf{93.3}      & \textbf{97.4}      & \textbf{98.5}      &    & \textbf{91.4}     &  \textbf{92.0}    &  \textbf{96.7}     &  \textbf{97.9}     \\ 
  \bottomrule
   \end{tabular}
   }
 \end{center} \vspace{-1em}
 \caption{Comparison of different association schemes upon our baseline method. Top-1.-5,-10 accuracies (\%) and mAP(\%) are reported.} \label{Comparison of variants} \vspace{-2em}
\end{table}

\subsection{The Effect of Global Discriminative Association (GDA)}

 \noindent \textbf{Comparison with non-discriminative variants.} We evaluate the effects of global discriminative image-language association by comparing the variants with and without using the description feature $\theta^{g}(T)$.  Among them, \emph{basel.+GDA} improves \emph{basel.} by 5.6\% and 4.4\% in term of mAP on Market-1501 and CUHK-SYSU respectively (Table. \ref{Comparison of variants}), which shows that GDA can benefit the learning of visual representation. Furthermore, our proposed method  yields better performance than $ \emph{basel.+LRA}$, indicating the effect of global discriminative association is complementary to that of the local reconstructive association.

 \begin{table}[b] \vspace{-1em}
\setlength{\tabcolsep}{6pt}
 \begin{center}
  \resizebox{0.95\hsize}{0.11\hsize}{
   \begin{tabular}{p{40pt}<{\centering}
                  p{3pt}<{\centering}
                  p{25pt}<{\centering}p{25pt}<{\centering}p{25pt}<{\centering}p{28pt}<{\centering}  
                  p{3pt}<{\centering}			
                  p{25pt}<{\centering}p{25pt}<{\centering}p{25pt}<{\centering}p{28pt}<{\centering}   
   }

    \toprule
    \emph{basel.+GDA}   &  & \multicolumn{4}{c}{Market-1501} &  & \multicolumn{4}{c}{CUHK-SYSU}  \\
                \cline{3-6}\cline{8-11}
     $ \lambda_{T}$  &    &  mAP     &  top-1      &  top-5     & top-10     &    &     mAP          & top-1   &  top-5  &  top-10 \\
   \hline
      0       &         &   78.9           &   91.1         &  96.4          &  97.6         &    &     89.3         &  90.3          &   95.7        &   97.0   \\
      0.05    &         &   79.2           &   91.2         &  96.5          &  97.8         &    &     89.2         &  90.2          &   95.6        &   96.9   \\
      0.1     &         &   \textbf{80.0}  &  \textbf{91.5} &  96.4          & \textbf{98.0} &    &    \textbf{90.2} &  \textbf{91.0} &  \textbf{96.3}&  \textbf{97.5}   \\
      0.5     &         &   79.6           &   91.3         &  96.6          &  97.9         &    &     89.8         &  90.9          &   96.2        &   97.3   \\   
       1      &         &   78.9           &   91.0         &  96.6          &  97.9         &    &     89.1         &  90.0          &   95.8        &   97.1   \\
\bottomrule
   \end{tabular}
   }   
 \end{center}
 \caption{ Importance analysis of $\mathcal{L}_{T}$ in  \emph{basel.+GDA}. We fix $\lambda_{dis}=1$ and adjust $\lambda_{T}$ over 0, 0.05, 0.1, 0.5, 1.  Top-1,-5,-10 accuracies (\%) and mAP(\%) are reported.} \label{Comparison of parameters}
\end{table}

  \noindent \textbf{Comparison with bi-directional ranking loss \cite{Dual-path, Karpathy:2014}.} $\mathcal{L}_{dis}$ in GDA aims to discriminate the matched image-text pairs from the unmatched ones. It has the similar functions  with the bidirectional ranking loss $\mathcal{L}_{rank}$ (Eqn. (\ref{Eq:loss_rank})) for image-language cross-modal retrieval.  We implement two types of ranking losses for comparison.  The first one is more similar to the loss in \cite{Karpathy:2014}, where a positive image-text pair is composed of the image and text from the same tuple. The other one adopts the loss in \cite{Dual-path}, where the positive image-text pairs are obtained by arbitrary image-text combinations from the same person.  We modify \emph{basel.+GDA} by replacing $\mathcal{L}_{dis}$ with the two loss functions,  and denote them by $\emph{basel.+rank}^{1} $ and $\emph{basel.+rank}^{2}$, respectively. The results in Table \ref{Comparison of variants} show that both ranking losses can boost the baseline. Besides, $\emph{basel.\!+\! rank}^{2} $ is better than $\emph{basel.\!+\! rank}^{1} $ by incorporating more abundant positive samples for discrimination. The proposed \emph{basel.\!+\! GDA} further improves the mAP by 2.3\% and 1.4\% on Market-1501 and CUHK-SYSU, verifying the effectiveness of our relevance estimation strategy (Eqns. \ref{Eqn:r1} and \ref{Eqn:r2} ).

\noindent   \textbf{The importance of $\mathcal{L}_{T}$.} To preserve separability of the visual feature, the associated linguistic feature $\theta^{g}(T)$ is supposed to be discriminative for different persons, thus $\mathcal{L}_{T}$ is employed along with $\mathcal{L}_{dis}$.  We investigate the importance of  $\mathcal{L}_{T}$ based upon \emph{basel.+GDA} and observe how the performance changes with $\lambda_{T}$ in Table \ref{Comparison of parameters}. Slightly worse results are observed when $\lambda_{T}=0$, indicating $\mathcal{L}_{T}$ is indispensable. On the other hand, the optimal results are achieved when $\lambda_{T}$ is around $0.1$. One possible reason is that language description is sometimes more ambiguous to describe a specific person, making $\mathcal{L}_{I}$ and $\mathcal{L}_{T}$ not equally important. For example, ``The man wears a blue shirt" can simultaneously describe different persons wearing a dark blue shirt and a light blue shirt.

\subsection{The Effect of Local Reconstructive Association (LRA)} 

\textbf{Comparison with non-reconstructive variants.} We evaluate the effects of local reconstructive association by comparing the variants with and without using the local phrase feature $\theta^{l}(P)$. The performance gap between \emph{basel.} and \emph{basel.+LRA} proves the effectiveness of LRA for visual feature learning.  Employing LRA brings $5.2\%$ and $3.9\%$  mAP gain over the two datasets, which is close to the gain of employing GDA.  Besides, the fact that the proposed method is better than \emph{basel.+GDA} also indicates the effectiveness of LRA.

\begin{figure}[t]
 \begin{center}
\includegraphics[width=0.8\textwidth]{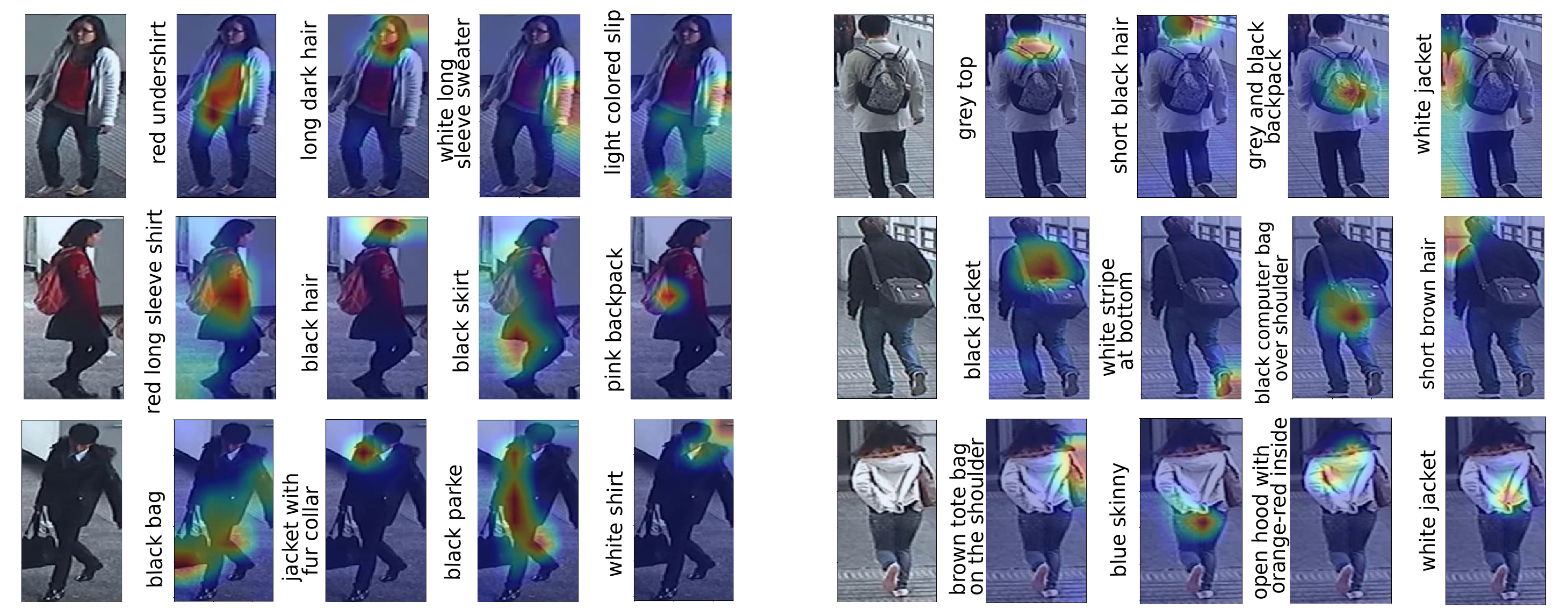}
 \vspace{-1em}
 \caption{Heat maps of the attention weights. The phrases are placed in on the left of the corresponding heat maps. Zoom in the figure for better view of the phrases.}  \vspace{-1em} \label{attention_masks} 
 \end{center}
\end{figure}

\noindent  \textbf{Visualization of phrase-guided attention weights.}  We compute the attention weights for a specific phrase (Eqn. (\ref{fig:local-reconstruction})), align the weights to the corresponding image, and obtain the heat map for the phrase. The heat maps are displayed in Fig. \ref{attention_masks}, showing that the attention weights can roughly capture the local regions described by the phrases.
  
\begin{table}[t]
   \begin{minipage}{0.75\textwidth}
  \resizebox{0.99\hsize}{0.075\hsize}{
   \begin{tabular}{p{66pt}<{\centering} p{25pt}<{\centering}p{25pt}<{\centering}p{28pt}<{\centering}       p{66pt}<{\centering}p{25pt}<{\centering}p{25pt}<{\centering}p{28pt}<{\centering}}
    \toprule
     Methods   &  top-1 	 &  top-5    &  top-10        & Methods  &     top-1 	 &  top-5    &  top-10  \\
\hline 
       GNA-RNN \cite{Lishuang_2017_CVPR}      &     19.05   &    - -    &   53.64    & DPCE \cite{Dual-path} &  \textbf{44.40}  &    66.26  &   75.07  \\   
       IATV  \cite{Li_2017_ICCV}        &     25.94    &    - -    &   60.49    & Ours      &  43.58      & \textbf{66.93}  &   \textbf{76.26} \\
  \bottomrule
   \end{tabular} }  
\end{minipage}
 ~\hfill~
\begin{minipage}{0.22\textwidth}
 \captionof{table}{Results on CUHK-PEDES. }   \label{cross-results}
\end{minipage} 
\vspace{-1em}
\end{table}

\subsection{Results on Text-to-image Retrieval}

\begin{figure}[t]
 \begin{center}
\includegraphics[width=0.98\textwidth]{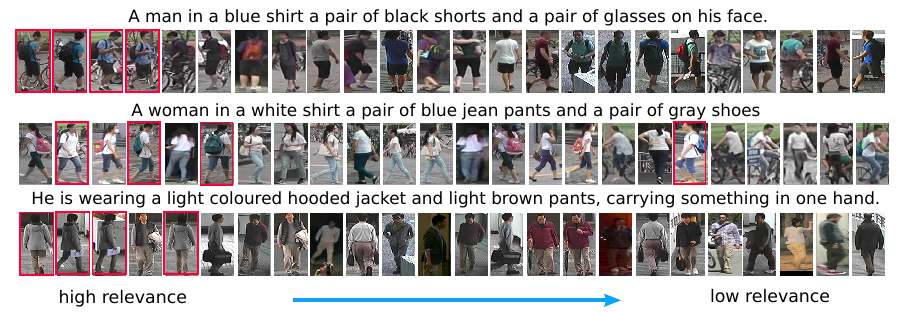}  \vspace{-1em} 
 \caption{Examples of the text-to-image search. The most relevant 24 images are displayed. Red boxes indicate the ground truth.}\label{fig:text-to-image-figure} 
 \end{center}\vspace{-2em}
\end{figure}

As a by-product, our method can also be utilized for text-to-image retrieval, which is fulfilled by ranking the cross-modal relevance (Eqn. (\ref{Eqn:r2})). We report the retrieval results on CUHK-PEDES following the standard protocol, where there are 3,074 test images with 6,156 captions, 3,078 validation images with 6,158 captions, and 34,054 training images with 68,126 captions. The quantitative and qualitative results are reported in Table \ref{cross-results} and Fig. \ref{fig:text-to-image-figure}, respectively. Although our method is not specifically designed for this task, it achieves competitive results to the current state-of-the-art methods. 

\begin{wraptable}{R}{60mm}
	\vspace{-1em}
   \resizebox{0.98\hsize}{0.35\hsize}{
   \begin{tabular}{p{112pt}
                  p{1pt}<{\centering}
                  p{25pt}<{\centering}p{25pt}<{\centering}p{28pt}<{\centering}}
      
    \toprule
     \multicolumn{1}{c}{\multirow{2}{*}{Methods}}   && \multicolumn{3}{c}{CUHK01}  \\
                  \cline{3-5}
               &      &   top-1 &  top-5  &  top-10    \\
   \hline 
     \cite{Liao_2017_CVPR_LOMO}   XQDA(CVPR15)           &      &   63.2     	 &    83.9    	    &   90.0     \\
     \cite{Xiao_2016_CVPR}   JSTL(CVPR16)           &      &   66.6     	 &    -   	        &   -         \\    
     \cite{Zhang_2016_CVPR_nullspace}  DNS(CVPR16)            &      &   69.1      	 &    86.9   	    &   91.8     \\
     \cite{Chen_2017_CVPR} Quad(CVPR17)       &      &   62.6          &    83.4          &   89.7     \\
     \cite{YingcongChen} CRAFT(PAMI17)              &      &   74.5          &    91.2          &   94.8     \\                      
     \cite{zhao2017spindle} Spindle(CVPR17)          &      &   79.9          &    94.4          &   97.1     \\
     \cite{Zhao_2017_ICCV} DLPAR (ICCV17)           &      &   75.0          &    93.5          &   95.5     \\
     \centering basel.                              &      &   77.0          &    93.2          &   95.3     \\
     \centering proposed                            &      &  \textbf{84.8}  &   \textbf{95.1}  &  \textbf{98.4}     \\               
  \bottomrule
   \end{tabular}}
 \captionof{table}{\small Results on CUHK01. Top-1,-5,-10 accuracies(\%) are reported.}  \vspace{-2em}  \label{Tab:CUHK01}
\end{wraptable}

\subsection{Comparison with the State-of-the-Art Approaches}

 We compare our method with the current state-of-the-arts on the Market1501, CUHK03, and CUHK01 datasets. The results on Market-1501 are reported in Table \ref{Tab:comparison1} \textbf{left}. Our method outperforms all the other approaches regarding mAP and top-1 accuracy under both single-query and multi-query protocols.  Note that the baseline of our method is quite competitive to the most of the previous methods, which is partly because of well initialized ResNet-50 backbone and  proper data augmentation strategies. The proposed image-language association scheme can largely boost the well-performed baseline, making our method better than the recent state-of-the-arts \cite{almazan2018re,deep-person}. CUHK03 has two types of person bounding boxes: one is manually labeled,  and the other is obtained by a pedestrian detector. We compare our methods and others on both types, and report the top-1 and top-5 accuracies in Table \ref{Tab:comparison1} \textbf{right}. It can be seen that our method has significant advantages over the  top-1 accuracy, but is 0.2\% less than D-person \cite{deep-person} on the top-5 accuracy for the labeled bounding boxes. As D-person only utilizes image data, it is promising to apply our language association scheme to D-person for better performance.  Compared with Market-1501 and CUHK03, CUHK01 has fewer images for training as described in Sec. \ref{Tab:CUHK01}.  As in Table \ref{Tab:CUHK01},
the proposed association schemes have 7.8\% top-1 accuracy gain over the baseline on CUHK01. The results confirm the effectiveness of language description, and indicate the schemes may be more useful when the image data are not enough. 
 
 Among the compared approaches, Spindle\cite{zhao2017spindle} and  PDC\cite{Su_2017_ICCV} utilize pose landmarks,  CADL\cite{Lin2017CVPRcamera} employs the camera ID labels, and ACN\cite{ACN} makes use of the attributes for training.
We achieve better results than them on all the three datasets (Tables \ref{Tab:CUHK01} and \ref{Tab:comparison1}).  The results indicate language description is also a kind of  useful auxiliary information for person re-ID. With the proposed schemes, it can achieve the superior performance with the standard CNN architecture.

 \begin{table}[t]
\setlength{\tabcolsep}{6pt}
 \begin{center}
  \resizebox{0.96\hsize}{0.28\hsize}{
   \begin{tabular}{p{103pt}
                  p{1pt}<{\centering}
                  p{23pt}<{\centering}p{23pt}<{\centering}p{23pt}<{\centering}p{23pt}<{\centering} 
                  p{1pt}<{\centering}			
                  p{23pt}<{\centering}p{23pt}<{\centering}p{23pt}<{\centering}p{23pt}<{\centering}  
   }
\toprule
   \multicolumn{1}{c}{\multirow{3}{*}{Methods}}           &      &        \multicolumn{4}{c}{Market-1501}                                   &      &      \multicolumn{4}{c}{CUHK03}  \\ 
                    \cline{3-6}\cline{8-11}
             
             &      & \multicolumn{2}{c}{Single Query} &     \multicolumn{2}{c}{Multi-Query}   &      & \multicolumn{2}{c}{Labeled}       &     \multicolumn{2}{c}{Detected} \\
                  \cline{3-11}
             &      &   mAP  &  top-1 &  mAP  &  top-1  &        &   top-1     &  top-5  &  top-1  &  top-5      \\
   \hline 
     \cite{Zhousan_2017_CVPR} P2S(CVPR17)       &&  44.3  & 70.7 &  55.7 & 85.8  &&   -    &  -   &  -    &  -   \\
     \cite{Lin2017CVPRcamera} CADL(CVPR17)                         &&  47.1  & 73.8 &  55.6 & 80.9  &&   -    &  -   &  -    &  -  \\
      \cite{Li_Danwei_2017_CVPR} MSCAN(CVPR17)        &&  57.5  & 80.3 &  66.7 & 86.8  &&  74.2  & 94.3 & 68.0  & 91.0 \\ 
      \cite{Bai_2017_CVPR}  SSM(CVPR17)          &&  68.8  & 82.2 &  76.2 & 88.2  &&  76.6  & 94.6 & 72.7  & 92.4 \\
      \cite{Zhong_2017_CVPR} k-rank(CVPR17)      &&  63.4  & 77.1 &  -    &  -    &&  61.6  &  -   & 58.5  &  -   \\
         \cite{ACN} ACN(CVPRW17)                       &&  62.6  & 83.6 &  -   & -    &&   -    &  -   &  62.6 &  89.7 \\
      \cite{Sun_2017_ICCV} SVDNet(ICCV17)        &&  62.1  & 82.3 &  -    &  -    &&   -    &  -   & 81.8  & 95.2 \\
      \cite{Zhao_2017_ICCV} DLPAR(ICCV17)           &&  63.4  & 81.0 &  -    &  -    &&  85.4  & 97.6 & 81.6  & 97.3 \\
      \cite{Zhou_2017_ICCV} OLMANS(ICCV17)       &&  60.7  &  -   & 66.8  &  -    &&  61.7  & 88.4 & 62.7  & 87.6 \\
     \cite{Qian_2017_ICCV} MuDeep(ICCV17)        &&   -    &  -   &  -    &  -    &&  76.9  & 76.3 & 75.6  & 94.4 \\             
      \cite{Su_2017_ICCV} PDC(ICCV17)            &&  63.4  & 84.1 &  -    &  -    &&  88.7  & 98.6 & 78.3  & 94.8 \\
     \cite{zheng2017unlabeled} VI+LSRO(ICCV17)   &&  66.1  & 84.0 & 76.1  & 88.4  &&   -    &  -   & 84.6  & 97.6 \\ 
  \cite{Chen_2017_ICCV_Workshops} DPFL(ICCVW17)  &&  73.1  & 88.9 & 80.7  & 92.3  &&  86.7  &  -   & 82.0  &  -   \\
      \cite{li2017person} JLMT(IJCAI17)          &&  65.5  & 85.1 & 74.5  & 89.7  &&  83.2  & 98.0  & 80.6  & 96.9 \\
      \cite{deep-person} D-Person(Arxiv17)       &&  79.6  & 92.3 & 94.5  & 85.1  &&  91.5  & \textbf{99.0}  & 89.4  & \textbf{98.2}  \\
      \cite{almazan2018re} TGP(Arxiv18)          &&  81.2  & 92.2 & 87.3  & 94.7  &&   -    &  -    &  -   &- \\
      \centering basel.                          &&  74.4  & 89.2 &  82.3  &  93.3     &&  88.4  & 98.1  & 87.9 & 97.5  \\
      \centering  proposed                       &&  \textbf{81.8} & \textbf{93.3} &  \textbf{87.9}  &  \textbf{95.3} &&  \textbf{92.5}  & 98.8  & \textbf{90.9} & \textbf{98.2} \\
  \bottomrule
   \end{tabular}
   }
   \end{center}
  \caption{Comparison with the state-of-the-art methods on the Market-1501 and CUHK03 datasets.  The results on Market-1501 are under single-query and multi-query protocols.  MAP (\%) and top-1 accuracy (\%) are reported.  Meanwhile, the performances on CUHK03 are evaluated with labeled and detected bounding boxes. Top-1 and Top-5 accuracies(\%) are reported.} \label{Tab:comparison1}  \vspace{-2em}
\end{table}

\section{Conclusions}

We utilized language descriptions as additional training supervisions to improve the visual features for person re-identification. The global and local image-language association schemes have been proposed. The former learns better global visual features with the discriminative supervision of the overall language descriptions, while the latter enforces the semantic consistencies between local visual features and noun phrases by phrase reconstruction. Our ablation studies show that the proposed image-language association schemes can remarkably improve the learning of the visual feature and are more effective than the existing image-text joint embedding methods. The proposed method achieves state-of-the-art performance on three public person re-ID datasets.

\vspace{1em}

\noindent \textbf{Acknowledgement}. \footnotesize{This work is supported by SenseTime Group Limited, the General Research Fund sponsored by the Research Grants Council of Hong Kong (Nos. CUHK14213616, CUHK14206114, CUHK14205615, CUHK14203015, CUHK14239816, CUHK419412, CUHK14207814, CUHK14208417, CUHK14202217), the Hong Kong Innovation and Technology Support Program (No.ITS/121/15FX).}

\bibliographystyle{splncs04}
\bibliography{lang}

\end{document}